\documentclass[runningheads]{llncs}
\usepackage{amsmath}
\usepackage{cancel}
\usepackage[linesnumbered,ruled,vlined]{algorithm2e}
\usepackage[T1]{fontenc}
\usepackage{graphicx,verbatim}
 
\usepackage{hyperref}
\usepackage{color}
\hypersetup{
  colorlinks=true,
  linkcolor=blue,
  citecolor=blue,
  urlcolor=blue
}

\usepackage{subcaption}

\begin{document}
\title{Mechanistic Learning with Guided Diffusion Models to Predict Spatio-Temporal Brain Tumor Growth}
%
\author{Daria Laslo\inst{1,2}\orcidID{0009-0005-8771-144X} \and
Efthymios Georgiou\inst{3}\orcidID{0000-0002-6042-9584} \and
Marius George Linguraru\inst{4}\orcidID{0000-0001-6175-8665}\and
Andreas M. Rauschecker \inst{5}\orcidID{0000-0003-0633-9876} \and
Sabine M\"uller \inst{5,6}\orcidID{0000-0002-3452-5150} \and
Catherine R. Jutzeler \inst{1,2}\orcidID{0000-0001-7167-8271} \and
Sarah Br\"uningk \inst{3}\orcidID{0000-0003-3176-1032}}

\authorrunning{D. Laslo et al.}
%
\institute{ETH Zurich, 8092 Zurich, Switzerland - 
\email{daria.laslo@hest.ethz.ch}\\
\and
SIB Swiss Institue of Bioinformatics, 1015 Lausanne, Switzerland
\and
Department of Radiation Oncology, Inselspital, Bern University Hospital and University of Bern, Switzerland\\
\and
Sheikh Zayed Institute for Pediatric Surgical Innovation, Children’s National Medical Center, Washington, DC 20010, United States
\and
University of California San Francisco, San Francisco, CA 94143, United States 
\and
University Children's Hospital Zurich, 8008 Zurich, Switzerland }


\newcommand{\head}[1]{{\smallskip\noindent\textbf{#1}}}
\newcommand{\alert}[1]{{\color{red}{#1}}}
\newcommand{\sm}{\scriptsize}
\newcommand{\eq}[1]{(\ref{eq:#1})}

\newcommand{\Th}[1]{\textsc{#1}}
\newcommand{\mr}[2]{\multirow{#1}{*}{#2}}
\newcommand{\mc}[2]{\multicolumn{#1}{c}{#2}}
\newcommand{\tb}[1]{\textbf{#1}}
\newcommand{\ch}{\checkmark}

\newcommand{\red}[1]{{\textcolor{red}{#1}}}
\newcommand{\blue}[1]{{\textcolor{blue}{#1}}}
\newcommand{\green}[1]{{\textcolor{green}{#1}}}
\newcommand{\gray}[1]{{\textcolor{gray}{#1}}}

\newcommand{\citeme}[1]{\alert{[X]}}
\newcommand{\refme}[1]{\alert{(X)}}

\newcommand{\fig}[2][1]{\includegraphics[width=#1\linewidth]{fig/#2}}
\newcommand{\figh}[2][1]{\includegraphics[height=#1\linewidth]{fig/#2}}
\newcommand{\figa}[2][1]{\includegraphics[width=#1]{fig/#2}}
\newcommand{\figah}[2][1]{\includegraphics[height=#1]{fig/#2}}

\newcommand{\tran}{^\top}
\newcommand{\mtran}{^{-\top}}
\newcommand{\zcol}{\mathbf{0}}
\newcommand{\zrow}{\zcol\tran}

\newcommand{\ind}{\mathbbm{1}}
\newcommand{\expect}{\mathbb{E}}
\newcommand{\nat}{\mathbb{N}}
\newcommand{\zahl}{\mathbb{Z}}
\newcommand{\real}{\mathbb{R}}
\newcommand{\proj}{\mathbb{P}}
\newcommand{\prob}{\operatorname{P}}
\newcommand{\normal}{\mathcal{N}}

\newcommand{\mif}{\textrm{if}\ }
\newcommand{\other}{\textrm{otherwise}}
\newcommand{\minimize}{\textrm{minimize}\ }
\newcommand{\maximize}{\textrm{maximize}\ }
\newcommand{\st}{\textrm{subject\ to}\ }

\newcommand{\id}{\operatorname{id}}
\newcommand{\const}{\operatorname{const}}
\newcommand{\sgn}{\operatorname{sgn}}
\newcommand{\var}{\operatorname{Var}}
\newcommand{\mean}{\operatorname{mean}}
\newcommand{\trace}{\operatorname{tr}}
\newcommand{\diag}{\operatorname{diag}}
\newcommand{\vect}{\operatorname{vec}}
\newcommand{\cov}{\operatorname{cov}}
\newcommand{\sign}{\operatorname{sign}}
\newcommand{\prj}{\operatorname{proj}}

\newcommand{\softmax}{\operatorname{softmax}}
\newcommand{\clip}{\operatorname{clip}}

\newcommand{\defn}{\mathrel{:=}}
\newcommand{\peq}{\mathrel{+\!=}}
\newcommand{\meq}{\mathrel{-\!=}}

\newcommand{\paren}[1]{\left({#1}\right)}
\newcommand{\set}[1]{\left\{{#1}\right\}}
\newcommand{\floor}[1]{\left\lfloor{#1}\right\rfloor}
\newcommand{\ceil}[1]{\left\lceil{#1}\right\rceil}
\newcommand{\inner}[1]{\left\langle{#1}\right\rangle}
\newcommand{\norm}[1]{\left\|{#1}\right\|}
\newcommand{\abs}[1]{\left|{#1}\right|}
\newcommand{\frob}[1]{\norm{#1}_F}
\newcommand{\card}[1]{\left|{#1}\right|\xspace}

\newcommand{\diff}{\mathrm{d}}
\newcommand{\der}[3][]{\frac{\diff^{#1}#2}{\diff#3^{#1}}}
\newcommand{\ider}[3][]{\diff^{#1}#2/\diff#3^{#1}}
\newcommand{\pder}[3][]{\frac{\partial^{#1}{#2}}{\partial{{#3}^{#1}}}}
\newcommand{\ipder}[3][]{\partial^{#1}{#2}/\partial{#3^{#1}}}
\newcommand{\dder}[3]{\frac{\partial^2{#1}}{\partial{#2}\partial{#3}}}

\newcommand{\wb}[1]{\overline{#1}}
\newcommand{\wt}[1]{\widetilde{#1}}

\def\xssp{\hspace{1pt}}
\def\ssp{\hspace{3pt}}
\def\msp{\hspace{5pt}}
\def\lsp{\hspace{12pt}}

\newcommand{\cA}{\mathcal{A}}
\newcommand{\cB}{\mathcal{B}}
\newcommand{\cC}{\mathcal{C}}
\newcommand{\cD}{\mathcal{D}}
\newcommand{\cE}{\mathcal{E}}
\newcommand{\cF}{\mathcal{F}}
\newcommand{\cG}{\mathcal{G}}
\newcommand{\cH}{\mathcal{H}}
\newcommand{\cI}{\mathcal{I}}
\newcommand{\cJ}{\mathcal{J}}
\newcommand{\cK}{\mathcal{K}}
\newcommand{\cL}{\mathcal{L}}
\newcommand{\cM}{\mathcal{M}}
\newcommand{\cN}{\mathcal{N}}
\newcommand{\cO}{\mathcal{O}}
\newcommand{\cP}{\mathcal{P}}
\newcommand{\cQ}{\mathcal{Q}}
\newcommand{\cR}{\mathcal{R}}
\newcommand{\cS}{\mathcal{S}}
\newcommand{\cT}{\mathcal{T}}
\newcommand{\cU}{\mathcal{U}}
\newcommand{\cV}{\mathcal{V}}
\newcommand{\cW}{\mathcal{W}}
\newcommand{\cX}{\mathcal{X}}
\newcommand{\cY}{\mathcal{Y}}
\newcommand{\cZ}{\mathcal{Z}}

\newcommand{\vA}{\mathbf{A}}
\newcommand{\vB}{\mathbf{B}}
\newcommand{\vC}{\mathbf{C}}
\newcommand{\vD}{\mathbf{D}}
\newcommand{\vE}{\mathbf{E}}
\newcommand{\vF}{\mathbf{F}}
\newcommand{\vG}{\mathbf{G}}
\newcommand{\vH}{\mathbf{H}}
\newcommand{\vI}{\mathbf{I}}
\newcommand{\vJ}{\mathbf{J}}
\newcommand{\vK}{\mathbf{K}}
\newcommand{\vL}{\mathbf{L}}
\newcommand{\vM}{\mathbf{M}}
\newcommand{\vN}{\mathbf{N}}
\newcommand{\vO}{\mathbf{O}}
\newcommand{\vP}{\mathbf{P}}
\newcommand{\vQ}{\mathbf{Q}}
\newcommand{\vR}{\mathbf{R}}
\newcommand{\vS}{\mathbf{S}}
\newcommand{\vT}{\mathbf{T}}
\newcommand{\vU}{\mathbf{U}}
\newcommand{\vV}{\mathbf{V}}
\newcommand{\vW}{\mathbf{W}}
\newcommand{\vX}{\mathbf{X}}
\newcommand{\vY}{\mathbf{Y}}
\newcommand{\vZ}{\mathbf{Z}}

\newcommand{\va}{\mathbf{a}}
\newcommand{\vb}{\mathbf{b}}
\newcommand{\vc}{\mathbf{c}}
\newcommand{\vd}{\mathbf{d}}
\newcommand{\ve}{\mathbf{e}}
\newcommand{\vf}{\mathbf{f}}
\newcommand{\vg}{\mathbf{g}}
\newcommand{\vh}{\mathbf{h}}
\newcommand{\vi}{\mathbf{i}}
\newcommand{\vj}{\mathbf{j}}
\newcommand{\vk}{\mathbf{k}}
\newcommand{\vl}{\mathbf{l}}
\newcommand{\vm}{\mathbf{m}}
\newcommand{\vn}{\mathbf{n}}
\newcommand{\vo}{\mathbf{o}}
\newcommand{\vp}{\mathbf{p}}
\newcommand{\vq}{\mathbf{q}}
\newcommand{\vr}{\mathbf{r}}
\newcommand{\Vs}{\mathbf{s}}
\newcommand{\vt}{\mathbf{t}}
\newcommand{\vu}{\mathbf{u}}
\newcommand{\vv}{\mathbf{v}}
\newcommand{\vw}{\mathbf{w}}
\newcommand{\vx}{\mathbf{x}}
\newcommand{\vy}{\mathbf{y}}
\newcommand{\vz}{\mathbf{z}}

\newcommand{\vone}{\mathbf{1}}
\newcommand{\vzero}{\mathbf{0}}

\newcommand{\valpha}{{\boldsymbol{\alpha}}}
\newcommand{\vbeta}{{\boldsymbol{\beta}}}
\newcommand{\vgamma}{{\boldsymbol{\gamma}}}
\newcommand{\vdelta}{{\boldsymbol{\delta}}}
\newcommand{\vepsilon}{{\boldsymbol{\epsilon}}}
\newcommand{\vzeta}{{\boldsymbol{\zeta}}}
\newcommand{\veta}{{\boldsymbol{\eta}}}
\newcommand{\vtheta}{{\boldsymbol{\theta}}}
\newcommand{\viota}{{\boldsymbol{\iota}}}
\newcommand{\vkappa}{{\boldsymbol{\kappa}}}
\newcommand{\vlambda}{{\boldsymbol{\lambda}}}
\newcommand{\vmu}{{\boldsymbol{\mu}}}
\newcommand{\vnu}{{\boldsymbol{\nu}}}
\newcommand{\vxi}{{\boldsymbol{\xi}}}
\newcommand{\vomikron}{{\boldsymbol{\omikron}}}
\newcommand{\vpi}{{\boldsymbol{\pi}}}
\newcommand{\vrho}{{\boldsymbol{\rho}}}
\newcommand{\vsigma}{{\boldsymbol{\sigma}}}
\newcommand{\vtau}{{\boldsymbol{\tau}}}
\newcommand{\vupsilon}{{\boldsymbol{\upsilon}}}
\newcommand{\vphi}{{\boldsymbol{\phi}}}
\newcommand{\vchi}{{\boldsymbol{\chi}}}
\newcommand{\vpsi}{{\boldsymbol{\psi}}}
\newcommand{\vomega}{{\boldsymbol{\omega}}}

\newcommand{\rLambda}{\mathrm{\Lambda}}
\newcommand{\rSigma}{\mathrm{\Sigma}}

\newcommand{\vLambda}{\bm{\rLambda}}
\newcommand{\vSigma}{\bm{\rSigma}}

\makeatletter
\newcommand*\bdot{\mathpalette\bdot@{.7}}
\newcommand*\bdot@[2]{\mathbin{\vcenter{\hbox{\scalebox{#2}{$\m@th#1\bullet$}}}}}
\makeatother

\makeatletter
\DeclareRobustCommand\onedot{\futurelet\@let@token\@onedot}
\def\@onedot{\ifx\@let@token.\else.\null\fi\xspace}

\def\eg{\emph{e.g}\onedot} \def\Eg{\emph{E.g}\onedot}
\def\ie{\emph{i.e}\onedot} \def\Ie{\emph{I.e}\onedot}
\def\cf{\emph{cf}\onedot} \def\Cf{\emph{Cf}\onedot}
\def\etc{\emph{etc}\onedot} \def\vs{\emph{vs}\onedot}
\def\wrt{w.r.t\onedot} \def\dof{d.o.f\onedot} \def\aka{a.k.a\onedot}
\def\iid{i.i.d\onedot} \def\wolog{w.l.o.g\onedot}
\def\etal{\emph{et al}\onedot}
\makeatother

\newcommand{\Bern}{\operatorname{Bern}}
\newcommand{\Beta}{\operatorname{Beta}}
\newcommand{\Dir}{\operatorname{Dir}}
\newcommand{\relu}{\operatorname{ReLU}}

\newcommand{\rtwo}{$R^2$}

\definecolor{Gray0}{gray}{0.4}
\definecolor{Gray1}{gray}{0.75}
\definecolor{Gray2}{gray}{0.80}
\definecolor{Gray2}{gray}{0.82}
\definecolor{Gray3}{RGB}{205, 246, 250} 
\definecolor{Gray4}{gray}{0.95}
\definecolor{my_Green}{RGB}{0,140,0}

\definecolor{LightBlue1}{RGB}{173, 216, 230} 
\definecolor{LightBlue2}{RGB}{135, 206, 250} 
\definecolor{LightBlue3}{RGB}{176, 224, 230} 

\newcommand{\efthygeo}[1]{\textcolor{red}{efthygeo: #1}}


    
\maketitle              
%

\begin{abstract}
Predicting the spatio-temporal progression of brain tumors is essential for guiding clinical decisions in neuro-oncology. We propose a hybrid \emph{mechanistic learning} framework that combines a mathematical tumor growth model with a guided denoising diffusion implicit model (DDIM) to synthesize anatomically feasible future MRIs from preceding scans. The mechanistic model, formulated as a system of ordinary differential equations, captures temporal tumor dynamics including radiotherapy effects and estimates future tumor burden. These estimates condition a gradient-guided DDIM, enabling image synthesis that aligns with both predicted growth and patient anatomy. We train our model on the BraTS adult and pediatric glioma datasets and evaluate on 60 axial slices of in-house longitudinal pediatric diffuse midline glioma (DMG) cases. Our framework generates realistic follow-up scans based on spatial similarity metrics. It also introduces \emph{tumor growth probability maps}, which capture both clinically relevant extent and directionality of tumor growth as shown by $95^{th}$ percentile Hausdorff Distance. The method enables biologically informed image generation in data-limited scenarios, offering generative-space-time predictions that account for mechanistic priors.



\end{abstract}
\section{Introduction}

Longitudinal imaging is a cornerstone of the clinical workflow in neuro-oncology. Quantitative tumor segmentations enable disease burden and response classification \cite{Erker2020-as,Wen2010-ex} in the longitudinal context. While mechanistic models of tumor volume dynamics have previously been explored \cite{Lipkova2019-hn,Bruningk2021-lz,Yankeelov2013-ht,Ezhov2019-gn}, they strongly compress the complex spatial and anatomical aspects of radiographic data. Spatio-temporal predictions that capture both the extent of the tumor and its anatomical location are clinically more informative.  

This is particularly critical for aggressive brain tumors located in sensitive regions, such as pediatric diffuse midline glioma (DMGs)\cite{Khalighi2024-ks}. Although generative modeling for brain tumor imaging has been conceptually established \cite{Peng2022-er,Pinaya2022-se,Wolleb2022-dx}, spatio-temporal generation of tumor growth remains underexplored, despite promising work in other neurological applications \cite{Litrico2024-wd}.

Denoising diffusion probabilistic models (DDPMs) \cite{Ho2020-wg}, known for their high-fidelity image synthesis \cite{Dhariwal2021-ok}, allow conditioning and guidance via external inputs \cite{Konz2024-oz,Ho2022-uw}. We leverage a guided denoising diffusion implicit model (DDIM) \cite{Song2020-vt,Wolleb2022-dx}, directed by a regressor’s gradient, to synthesize future scans reflecting increased tumor burden.

However, this straightforward approach lacks the temporal component, remaining conditioned on static targets (i.e., tumor burden). An additional challenge is the scarcity of extensive longitudinal data needed to train spatio-temporal generative models, particularly in rare and fatal diseases such as DMG.

In this work, we propose a mechanistic learning \cite{Metzcar2024-it} approach that offers spatio-temporal brain tumor growth predictions given sparse and irregular temporal data. By integrating a mechanistic ordinary differential equation (ODEs) model of tumor dynamics with guided DDIMs, we are able to generate high fidelity multimodal MRI images. We evaluate image quality and prediction performance, demonstrating that our generated images conform with observed tumor growth patterns while preserving important anatomical structures. Our framework
enables biologically informed generation of future tumor states, supporting anticipatory symptom management and therapy planning.

\section{Methods}
\subsection{Guided Denoising Diffusion}

DDPM's forward process is based on small amounts of noise added iteratively for $L$ time steps to an input image \(x_0\), obtaining a series of increasingly noisy images: $x_1, x_2, \ldots, x_L$. Generation is achieved by learning the reverse diffusion process, which iteratively denoises the corrupted image by obtaining $x_{l-1}$ from $x_l$ step by step until recovering $x_0$. Mathematically, this translates to learning the conditional distribution $p(x_{l-1}|x_l)$. A U-Net-based architecture ($\epsilon_\theta$) \cite{Ronneberger2015-se} is trained to approximate this distribution through a parametrized conditional distribution $p_\theta(x_{l-1}|x_l)$. Following the DDPM formulation \cite{Ho2020-wg,Song2020-vt}, $x_{l-1}$ can be generated from $x_{l}$ via:
\begin{equation}\label{eq:xt-1final}
x_{l-1} = \sqrt{\bar{\alpha}_{l-1}}\left(\frac{x_l - \sqrt{1-\bar{\alpha}_l} \epsilon_\theta^{(l)}(x_l)}{\sqrt{\bar{\alpha}_l}}\right) + \sqrt{1-\bar{\alpha}_{l-1}-\sigma_l^2} \epsilon^{(l)}_\theta(x_l) + \sigma_l \epsilon_l
\end{equation}
Setting $\sigma_l = 0$ results in a deterministic generative process, known as DDIM \cite{Song2020-vt}. In our framework, we use the DDIM variant of our trained DDPM during inference to enable fast and reproducible image generation.

To synthesize axial brain slices with increased tumor burden, we follow the guidance approach from \cite{Dhariwal2021-ok}, training a separate regressor R to predict tumor size relative to brain area using images with varying noise levels (from 0 to $L$ diffusion steps). Following \cite{Wolleb2022-dx}, the regressor’s gradient, scaled by a parameter $s_{R}$, directs the generation process toward images with the desired tumor size. The influence of this guidance is controlled by a weighting parameter $s_{R}$ (regressor scale), which scales the gradient term, as described in the equation below.

\begin{equation}
\overline{\epsilon}^{(l)}_\theta\left(x_l\right)=\epsilon^{(l)}_\theta\left(x_l\right)-s_{R}\sqrt{1-\bar{\alpha}_l} \nabla_{x_l} R(x_{l}, l)
\end{equation}
The regressor scale $s_{R}$ can be modulated in two ways: dynamically based on the difference between the current regressor output and the target size, and statically via a scaling factor. Formally, we introduce $s_{R}=s_{R(ct)}*s_{R(dyn)}$, where $s_{R(ct)}$ is a constant scaling factor, and $s_{R(dyn)}$ is adaptively updated during sampling. As such, our framework enables the synthesis of brain MRIs of predefined tumor sizes, which we further combine with a mechanistic tumor growth model to simulate biologically plausible progression.

\subsection{Mechanistic Model}
We developed a mechanistic ODE model to characterize tumor growth dynamics in response to radiotherapy (RT). The model captures both intrinsic tumor growth and the delayed effects of RT-induced cell death, incorporating biologically motivated transitions in tumor composition\cite{Zheng2025-ta}.

\subsubsection{Tumor Growth Prior to RT.} Before the initiation of RT, tumor growth is assumed to follow exponential kinetics; a standard assumption for early-stage, untreated tumors, where resource limitations (e.g., oxygen, nutrients) have not yet induced growth saturation:
\begin{equation}
A(t) = A_0 e^{\lambda t}, \quad t < t_{\text{RTstart}}
\end{equation}
where \( A_0 \) is the initial axial area and \( \lambda \) is the net growth rate. 

\subsubsection{Compartmental Partitioning at RT Onset.} At the onset of RT (\( t = t_{\text{RTstart}} \)), the tumor is partitioned into two subpopulations to reflect differential cellular fates following irradiation: a surviving fraction (\( A_l \)) and a dying fraction (\( A_d \)):
\begin{equation}
A_l(t_{\text{RTstart}}) = S \cdot A(t_{\text{RTstart}}), \quad A_d(t_{\text{RTstart}}) = (1 - S) \cdot A(t_{\text{RTstart}})
\end{equation}
The parameter \( S \in [0, 1] \) represents the fraction of clonogenically surviving tumor and is conceptually analogous to surviving fraction metrics from radiobiology. This compartmentalization enables modeling of delayed cell death by mitotic catastrophe\cite{Hazout2025-bk}.

\subsubsection{Post-RT Dynamics.}
Following RT, the surviving population continues to proliferate exponentially:
\begin{equation}
A_l(t) = A_l(t_{\text{RTstart}}) e^{\lambda (t - t_{\text{RTstart}})}
\end{equation}
\( A_d \), representing cells that have sustained lethal DNA damage but may persist transiently, undergoes delayed shrinkage governed by a time-dependent decay rate:
\begin{equation}
A_d(t) = A_d(t_{\text{RTstart}}) e^{\lambda'(t) (t - t_{\text{RTstart}})}
\end{equation}

where $\lambda'(t) = \lambda_{\text{decay}} \tanh\left((t - t_{\text{RTstart}} - \delta) \cdot \text{slope} \right)$.
The decay rate \( \lambda'(t) \) captures the delayed onset and time course of RT-induced cell death. The $\tanh$ function enables a smooth transition from exponential growth to exponential decay at rate \( \lambda_{\text{decay}} \), modulated by a delay parameter \( \delta \) and a slope parameter controlling transition steepness. This form was chosen to reflect the gradual engagement of apoptotic and mitotic catastrophe pathways observed in irradiated tumors. The total tumor area for \( t \geq t_{\text{RTstart}} \) is given by the sum of the two compartments $A(t) = A_l(t) + A_d(t)$.

This model structure enables physiologically interpretable parameters while capturing essential features of tumor response to RT, including subpopulation dynamics, delayed shrinkage, and regrowth potential.

\begin{algorithm}[H]
\footnotesize
\DontPrintSemicolon
\SetAlgoLined
\KwIn{$A_0$, $\lambda$, delay, slope, $t$, $S$ (survival fraction)}
\KwOut{$A(t)$ (tumor area/volume)}

\textbf{Part 1: Compute Tumor Growth $A(t)$}\;

\uIf{$t < t_{\text{RTstart}}$}{
    $A(t) \leftarrow A_0 \cdot e^{\lambda t}$\;
}
\Else{
    $A_{l}(t_{\text{RTstart}}) \leftarrow S \cdot A(t_{\text{RTstart}})$\;
    $A_{d}(t_{\text{RTstart}}) \leftarrow (1 - S) \cdot A(t_{\text{RTstart}})$\;
    
    $\lambda_{\text{eff}} \leftarrow -\lambda \cdot \tanh\left(\frac{(t - t_{\text{RTstart}} - \text{delay})}{\text{slope}}\right)$\;
    
    $A_{l}(t) \leftarrow A_{l}(t_{\text{RTstart}}) \cdot e^{\lambda (t - t_{\text{RTstart}})}$\;
    $A_{d}(t) \leftarrow A_{d}(t_{\text{RTstart}}) \cdot e^{\lambda_{\text{eff}} (t - t_{\text{RTstart}})}$\;
    
    $A(t) \leftarrow A_l(t) + A_d(t)$\;
}
\textbf{Part 2: Bootstrap Uncertainty Quantification}\;
\For{$i \leftarrow 1$ \KwTo $N_{\text{bootstrap}}$}{
    $\tilde{A}_i \leftarrow A_{\text{observed}} + \mathcal{N}(0, \sigma_{\text{noise}}^2)$\;
    
    $\theta_i \leftarrow \arg\min_{\theta} \sum_{t} \left|\tilde{A}_i(t) - A(t; \theta)\right|^2$\;
    
    $A_{\text{pred},i}(t) \leftarrow A(t; \theta_i)$\;
}

$A_{\text{median}}(t) \leftarrow \text{median}\{A_{\text{pred},i}(t)\}_{i=1}^{N_{\text{bootstrap}}}$\;
$A_{\text{CI}}(t) \leftarrow \text{percentile}\{A_{\text{pred},i}(t)\}_{i=1}^{N_{\text{bootstrap}}}$ at $[2.5\%, 97.5\%]$\;

\Return $A_{\text{median}}(t)$, $A_{\text{CI}}(t)$\;
\caption{Tumor Growth Modeling \& Confidence Quantification}
\label{alg:tumor_growth}
\end{algorithm}

\subsection{Inference using Mechanistic Learning}

At inference time, we integrate the mechanistic tumor growth model into the guided diffusion framework to generate follow-up MRI scans. Given a series of tumor measurements, we fit a patient-specific model that allows temporal extrapolation of tumor burden. This model provides an estimate of the expected tumor size at the desired follow-up time.

To generate the follow-up scan, we start from the most recent reference image (e.g., from the last visit), which is noised by applying a fixed number of forward diffusion steps corresponding to a chosen noise level (NL). During the denoising (reverse diffusion) process, the extrapolated tumor size from the mechanistic model serves as a target for the regressor-based gradient guidance, allowing the model to synthesize an image consistent with both the input anatomy and the expected tumor progression. Specifically, we use the mechanistic model estimation to dynamically adjust the strength of the gradient ($\nabla_{x_l} R(x_{l},l)$) based on the current evaluation compared to the set target through $s_{R(dyn)}$, that gets amplified by the constant scaling $s_{R(ct)}$ to define the final regressor scale $s_{R}$.

\begin{figure}[h!]
\includegraphics[width=\textwidth]{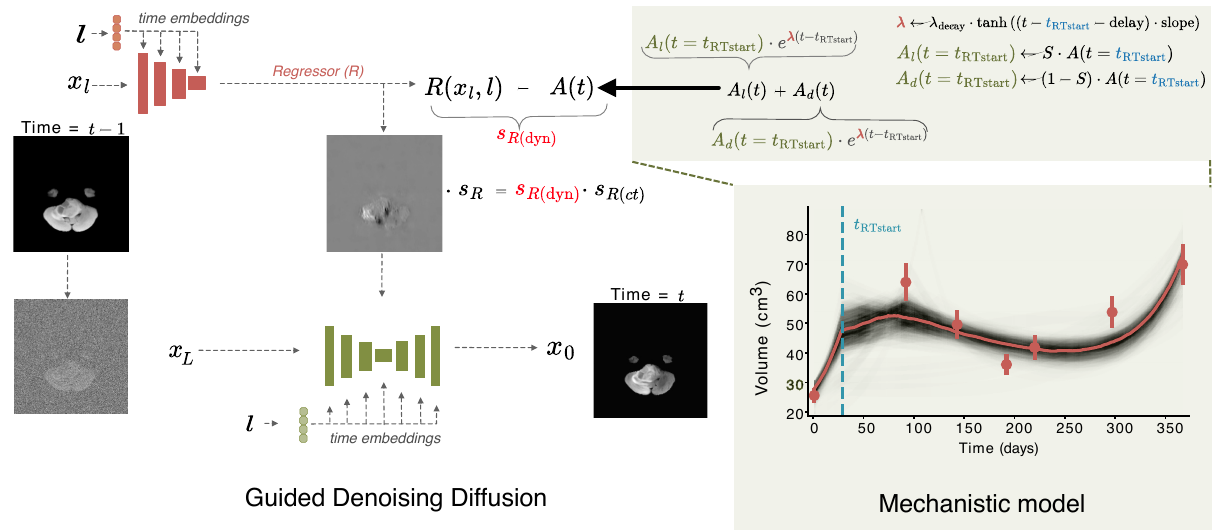}
\caption{Overview of the proposed method.} \label{fig1}
\end{figure}
\section{Experimental Results}
\subsection{Optimization of Guided Denoising Diffusion Model}
\subsubsection{Implementation Details.} 
For training our guided diffusion framework, we used multiparametric MRI scans from 1125 adult and 105 pediatric high-grade glioma patients from BraTS 2023 Challenge\cite{Menze2015-gq,Fathi-Kazerooni2025-am}. All 2D axial slices containing brain tissue were included (n$\sim$140,000). The DDPM was trained with 1,000 diffusion steps on the full dataset using a hybrid loss\cite{Nichol2021-sc} over 175,000 training steps, with Adam optimizer. Independently, a regressor was trained on the same data to predict tumor size relative to brain volume, using 60,000 training steps, with Adam optimizer and mean squared error (MSE) loss. A validation set comprising 26 adult and 9 pediatric cases was used to compute validation loss and determine early stopping. 

To evaluate inference performance, a set of longitudinally paired images (n=185) from 29 pediatric DMG patients (before/during RT) collected from DMG Center Zurich was used. Ethical approval for this study was granted based on 2022-00312 by the North-West and Central Swiss Ethics Commission. Images were processed using the established BraTS pipeline \cite{Pati2020-py}, longitudinally co-registered, and manually reviewed based on automated tumor segmentation. For the optimization of $\textit{NL}$ and $s_{R(ct)}$, we set the target tumor size as known at the next imaging session. A grid search over a predefined parameter space was conducted. 
\begin{figure}
\includegraphics[width=\textwidth]{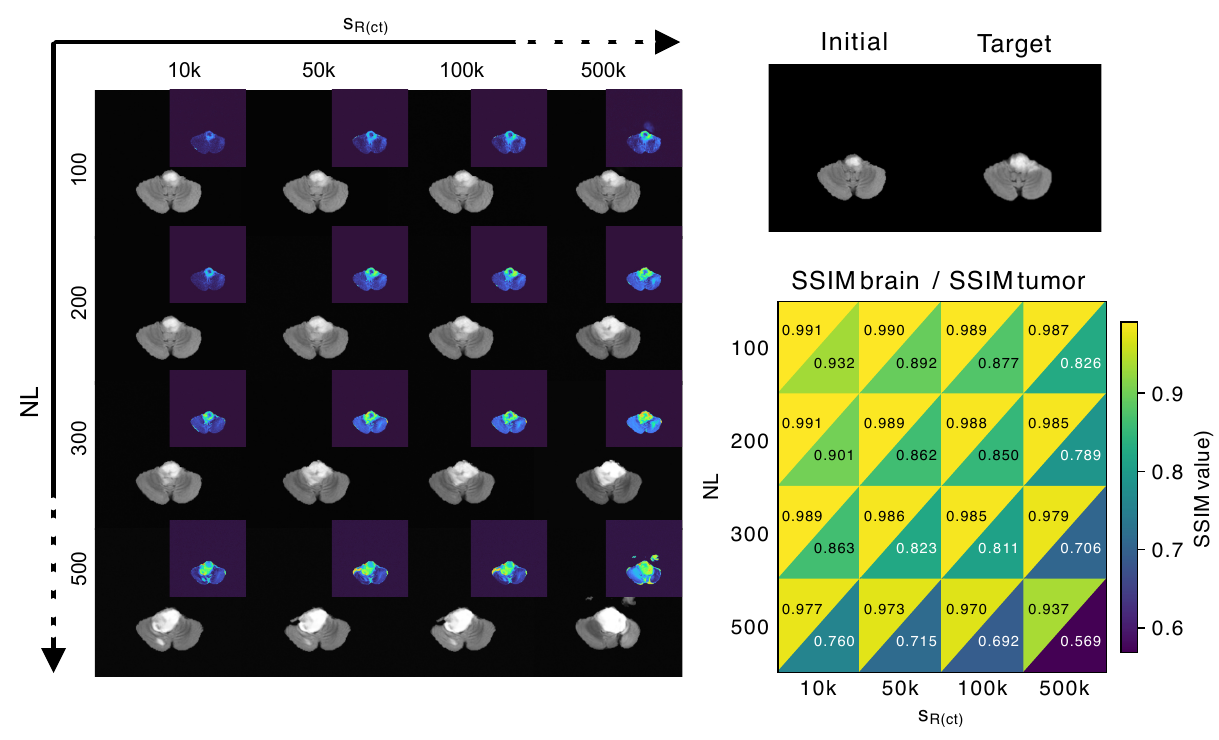}
\caption{Result for the grid search optimization of noise level (NL) and regressor scale ($s_{R(ct)}$). Generated images are shown along their difference maps to the starting image.} 
\label{fig1}
\end{figure}
\subsubsection{Evaluation.}
To assess the quality of generated images, we computed the Structural Similarity Index (SSIM) to the input image, separately within the tumor region and outside the extended tumor region. The regions were defined based on a dilated version (kernel adjusted to enable doubling of original tumor area) of the manual ground truth segmentation to account for the generated growth area. The region-specific evaluation was essential to capture focal changes (tumor) and the global consistency of the brain structure. Quantitative evaluation was complemented by qualitative visual assessments of the generated images compared to the actual follow-up scan, focusing on tumor progression patterns, anatomical feasibility, and the presence of artifacts or hallucinated structures.
\subsubsection{Results.}
Fig.\ref{fig1} shows the generated images given a grid of $NL$ and $s_{R(ct)}$ values, along corresponding difference maps comparing them to the input image (initial). Increasing $NL$ leads to greater alterations in brain anatomy, with hallucinated structures appearing at the highest value ($NL=500$). This is expected as a higher number of diffusion steps enables more flexibility in image synthesis.Similarly, increasing $s_{R(ct)}$ intensifies the tumor growth area, consistent with its radiological appearance on T2-weighted FLAIR, but excessive values ($\geq100k$) produce unrealistic hyperintensities. The effects between the two parameters are cumulative, highlighted for extreme values ($s_{R(ct)}=500k$ and $NL=500$). The trends are consistent across the dataset as shown by the average region-specific SSIM values in Fig.\ref{fig1}. Based on both quantitative metrics and visual assessments of the generated images for the longitudinal pairs, we selected  $NL=200$ and $s_{R(ct)}=50k$ for the downstream mechanistic learning predictions. 
%
%
\subsection{Mechanistic Modeling}
\subsubsection{Implementation Details.}
Our ODE model was fitted to longitudinal data from 8 pediatric DMG patients (60 slices) excluded from guided diffusion training, collected from DMG Center Zurich. Five equally spaced axial tumor slices, including the cnetral one, were selected from the initial and largest tumor burden imaging sessions if a growth more than $10\%$ was observed, totaling 60 2D axial slices treated independently. This patient and slice-specific approach captures individual tumor growth patterns and spatial heterogeneity.
All implementations used \texttt{SciPy}'s~\cite{2020SciPy-NMeth} \texttt{odeint} for numerical solving across \textit{pre} and \textit{post} RT phases. Parameter estimation used the \texttt{lmfit} library\footnote{with nonlinear least-squares optimization} to minimize residuals between simulated and observed tumor trajectories. We performed 100 bootstrap iterations with $10\%$ multiplicative Gaussian noise perturbation on tumor measurements. The model was re-fitted independently for each bootstrap replicate using different parameter constraints.

\subsubsection{Evaluation.}
We evaluated our mechanistic model using two experimental setups: \textit{all} used all available time-series data points to assess model accuracy in reproducing observed tumor growth patterns, while \textit{train} used all points except the last, with the final time point reserved for testing predictive ability. The predictive horizon of our split is between $[$36, 141$]$ days with a median of 67.

Performance was assessed using: 1) the coefficient of determination (\rtwo) to measure variance explained by our model, calculated between median predictions (over all bootstraps) and ground truth tumor area values and 2) normalized root mean square error (nRMSE) to quantify tumor size prediction error in the \textit{train} setup, calculated between the median predicted value (over all bootstraps) and the ground truth at the last (unseen) timestep. RMSE was normalized by dividing by the ground truth reference tumor area value. 

We selected the ODE fits with the best $R^2$ score on the \textit{train} setup, as it represents the realistic clinical scenario where we have access to all measurements except the last and want to predict tumor area at the next time point.

\begin{figure}[h!]
\includegraphics[width=\textwidth]{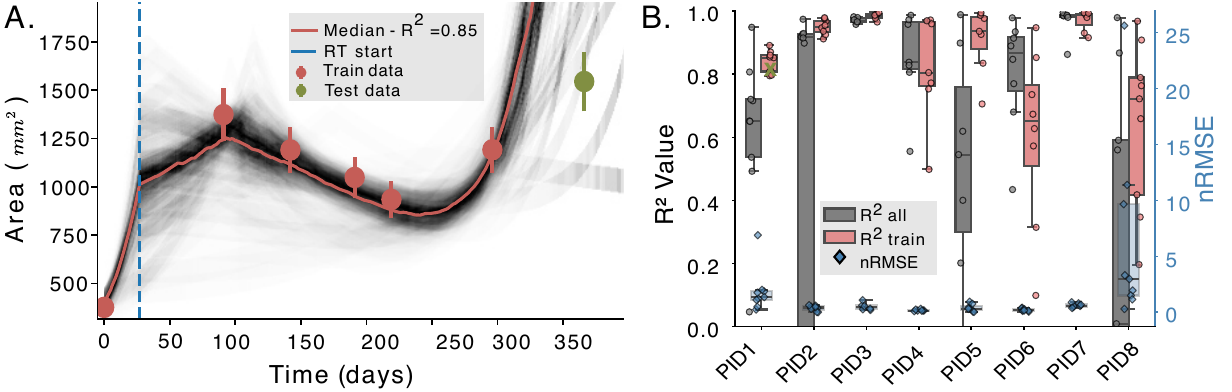}
\caption{A.Growth curve fitting (red) and estimation(gray). B. Mechanistic modeling performance metrics. Green cross highlights the example displayed in A. } \label{fig_MM_perf}
\end{figure}
\subsubsection{Results.}
Fig.~\ref{fig_MM_perf}A presents a representative example of 100 fitted tumor area trajectories (bootstraps) for the \textit{train} setup with the ground truth tumor area of the last time point (excluded from fitting) shown for reference. We observe that our ODE model solutions accurately capture tumor growth dynamics. Fig.~\ref{fig_MM_perf}B summarizes the fitting and predictive performance of the ODE model. The \textit{all} setup shows larger variance in $R^2$ values compared to the \textit{train} setup. We focus on the \textit{train} setup as it is clinically more relevant. The results demonstrate median $R^2$ values above 0.6 for the \textit{train} setup, confirming that our ODE models effectively approximate tumor growth dynamics across patients and slices.

We further evaluate the predictive ability of these models through normalized RMSE (nRMSE) in Fig.~\ref{fig_MM_perf}B. The results show consistently low median nRMSE values (median over 60 slices is 0.468) with small variance across patients and slices, indicating reliable tumor area forecasts. This robust predictive performance, combined with good fitting ability, confirms that our ODE model effectively captures DMG growth patterns and provides an appropriate framework for modeling tumor dynamics in this data-sparse pediatric cancer.

\subsection{Longitudinal Validation using Mechanistic Learning}
\subsubsection{Implementation Details.} We applied the integrated mechanistic learning framework to 60 2D axial brain slices from 8 additional pediatric patients (see \textit{Mechanistic Modeling} subsection). Previous available tumor area measurements were used to parametrize the ODE model as described. The tumor sizes below $90^\text{th}$ percentile of the bootstrap-estimated (n=100) values were used as targets to simulate plausible tumor growth trajectories and their anatomical evolution. Tumor growth probability maps are computed as an average of difference maps between the generated and original tumor MRI scans that were previously binarized using the Otsu method. We refer to these as dynamic probability maps as they aggregate values from generations using varying target sizes. As a comparison, the framework was applied with the target set to the true tumor size. In this case, generations were repeated using randomly sampled noise for the forward diffusion process, enabling the aggregation of the resulting images as a static (same target size) tumor growth probability map.   
\subsubsection{Evaluation.}
We thresholded both the generated static (true tumor size) and dynamic (ODE-estimated tumor size) probability maps to generate binary masks. These were compared with the true target segmentation using the $95^\text{th}$ percentile of the Hausdorff distance (HD95). For each slice, the HD95 was also computed between the initial and target image as a reference. A Wilcoxon-signed rank test was used to assess significance of HD95 change.
\subsubsection{Results.}
\begin{figure}[h!]
\includegraphics[width=\textwidth]{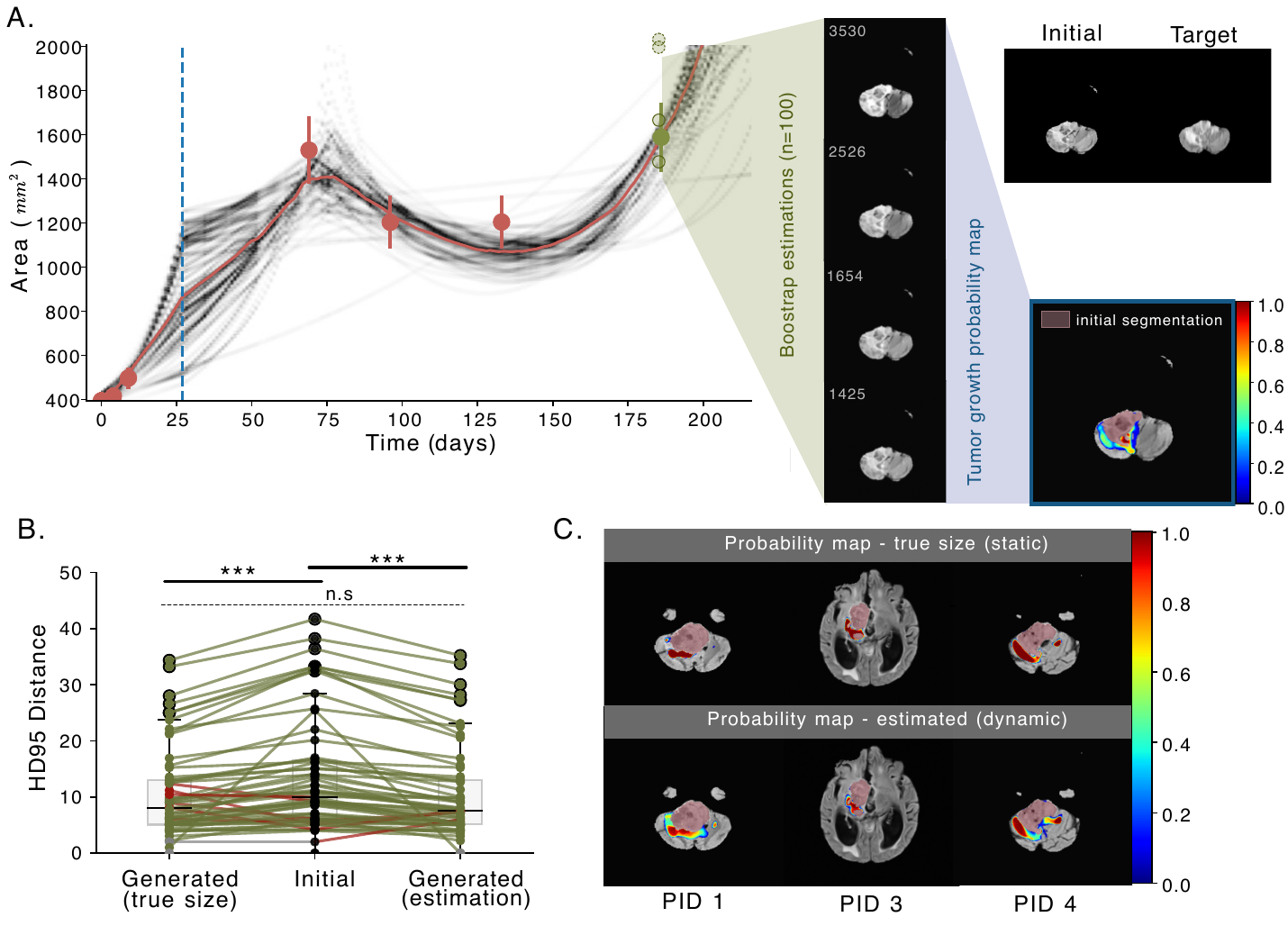}
\caption{A. Generating tumor growth probability maps based on boostrap-estimated tumor sizes. B. Quantitative evaluation. C. Tumor probability map comparison.} \label{fig_long}
\end{figure}
Fig.\ref{fig_long}A highlights T2-FLAIR images generated towards having tumor sizes as sampled from the bootstrap distribution, illustrating varying degrees of tumor progression. This progression is mirrored in the aggregated probability maps, where high-probability regions (shown in red) correspond closely to the actual areas of tumor growth observed in the target images. Notably, the smooth spatial transitions across probability bands indicate consistent growth trajectories across tumor size variations, highlighting the robustness of the generative model. Quantitative evaluation in (Fig\ref{fig_long}B, C) supports that the HD95 distance between the generated mask and true targets is significantly lower ($p_\text{value}\leq0.0005$) (green) than that between the initial image and the target. This supports the ability of the proposed framework to generate anatomically feasible tumor growth in the correct direction. Notably, the HD95 variability is higher in the case of thresholded static probability maps than when using the dynamic ones. This may reflect differences in tumor dynamics across samples, driven by variation in initial tumor sizes and the corresponding growth, which is factored in the dynamic maps through the bootstrap sampling covering a wider range of possible changes.

\section{Conclusion}
This work introduces a mechanistic learning framework for generative spatio-temporal tumor growth modeling, tailored towards data-sparse and unstructured scenarios. Our framework couples mechanistic ODE tumor growth models with generative models to construct a biologically informed image generation pipeline.

The mechanistic module, implemented via a biologically plausible ODE system, predicts the tumor area growth over time per patient and per slice. By feeding the mechanistic model's estimations to the diffusion model, we generate high fidelity brain tumor MRI scans conditioned on a target tumor area. Importantly, we aggregated generated images obtained from bootstrap estimates and synthesized both follow-up scans and tumor growth probability maps, which we showed to capture the directionality of tumor growth, making them highly valuable in the clinic. 

Despite these promising results, some limitations remain. The mechanistic model assumes access to at least a minimal number of imaging time points to accurately capture tumor growth trajectories, while the guided diffusion network relies on using multiparametric MRI scans as input. Generalization could be improved through the application of the proposed methodology to single contrast which should be assessed via an ablation study. Moreover, future work could explore patient-specific fine-tuning of the guided diffusion process, including generating tumors of varying sizes and optimizing structural similarity (SSIM) between real and synthesized brain regions and tumors.

Taken together, our experiments evaluated each module of the framework separately and in combination, across different clinical setups, demonstrating the feasibility of such a hybrid approach. Our synergistic mechanistic learning framework shows promising results which pave a promising direction for biologically informed longitudinal tumor growth image generation. Our experiments illustrate the direct application of our framework in pediatric DMG patients.


    

\begin{credits}
\subsubsection{\ackname}  
This study was funded by Olga Mayenfisch Foundation, SNF (CRSK-3\_229103, TMSGI3\_225913), Chad Though Defeat Cancer Foundation.
\subsubsection{\discintname}
The authors have no competing interests to declare that are
relevant to the content of this article. 
\end{credits}

\subsubsection{Code availability} 
The source code is available at:
 \href{https://github.com/CAIROLab-Bern/Mechanistic-Learning-Brain-Tumor-Growth}{https://github.com/CAIROLab-Bern/Mechanistic-Learning-Brain-Tumor-Growth}.
%
%
%
\bibliographystyle{splncs04}
\bibliography{references}

\begin{thebibliography}{10}
\providecommand{\url}[1]{\texttt{#1}}
\providecommand{\urlprefix}{URL }
\providecommand{\doi}[1]{https://doi.org/#1}

\bibitem{Bruningk2021-lz}
Brüningk, S.C., Peacock, J., Whelan, C.J., Brady-Nicholls, R., Yu, H.H.M., Sahebjam, S., Enderling, H.: Intermittent radiotherapy as alternative treatment for recurrent high grade glioma: a modeling study based on longitudinal tumor measurements. Sci. Rep.  \textbf{11}(1),  20219 (12~Oct 2021). \doi{10.1038/s41598-021-99507-2}

\bibitem{Dhariwal2021-ok}
Dhariwal, P., Nichol, A.: Diffusion models beat gans on image synthesis. Advances in neural information processing systems  \textbf{34},  8780--8794 (2021)

\bibitem{Erker2020-as}
Erker, C., Tamrazi, B., Poussaint, T.Y., Mueller, S., Mata-Mbemba, D., Franceschi, E., Brandes, A.A., Rao, A., Haworth, K.B., Wen, P.Y., Goldman, S., Vezina, G., MacDonald, T.J., Dunkel, I.J., Morgan, P.S., Jaspan, T., Prados, M.D., Warren, K.E.: Response assessment in paediatric high-grade glioma: recommendations from the response assessment in pediatric neuro-oncology ({RAPNO}) working group. Lancet Oncol.  \textbf{21}(6),  e317--e329 (Jun 2020). \doi{10.1016/S1470-2045(20)30173-X}

\bibitem{Ezhov2019-gn}
Ezhov, I., Lipkova, J., Shit, S., Kofler, F., Collomb, N., Lemasson, B., Barbier, E., Menze, B.: Neural parameters estimation for brain tumor growth modeling. arXiv [q-bio.QM]  (1~Jul 2019)

\bibitem{Fathi-Kazerooni2025-am}
Fathi~Kazerooni, A., Khalili, N., Liu, X., Haldar, D., Jiang, Z., Zapaishchykova, A., et~al.: {BraTS}-{PEDs}: Results of the multi-consortium international pediatric brain tumor segmentation challenge 2023. J. Mach. Learn. Biomed. Imaging  \textbf{3}(June 2025),  72--87 (26~Jun 2025). \doi{10.59275/j.melba.2025-f6fg}

\bibitem{Hazout2025-bk}
Hazout, S., Oehler, C., Zwahlen, D.R., Taussky, D.: Historical view of the effects of radiation on cancer cells. Oncol. Rev.  \textbf{19},  1527742 (30~Apr 2025). \doi{10.3389/or.2025.1527742}

\bibitem{Ho2020-wg}
Ho, J., Jain, A., Abbeel, P.: Denoising diffusion probabilistic models. Advances in neural information processing systems  \textbf{33},  6840--6851 (2020)

\bibitem{Ho2022-uw}
Ho, J., Salimans, T.: Classifier-free diffusion guidance. arXiv [cs.LG]  (25~Jul 2022)

\bibitem{Khalighi2024-ks}
Khalighi, S., Reddy, K., Midya, A., Pandav, K.B., Madabhushi, A., Abedalthagafi, M.: Artificial intelligence in neuro-oncology: advances and challenges in brain tumor diagnosis, prognosis, and precision treatment. NPJ Precis. Oncol.  \textbf{8}(1), ~80 (29~Mar 2024). \doi{10.1038/s41698-024-00575-0}

\bibitem{Konz2024-oz}
Konz, N., Chen, Y., Dong, H., Mazurowski, M.A.: Anatomically-controllable medical image generation with segmentation-guided diffusion models. arXiv [eess.IV]  (7~Feb 2024)

\bibitem{Lipkova2019-hn}
Lipkova, J., Angelikopoulos, P., Wu, S., Alberts, E., Wiestler, B., Diehl, C., Preibisch, C., Pyka, T., Combs, S.E., Hadjidoukas, P., Van~Leemput, K., Koumoutsakos, P., Lowengrub, J., Menze, B.: Personalized radiotherapy design for glioblastoma: Integrating mathematical tumor models, multimodal scans, and bayesian inference. IEEE Trans. Med. Imaging  \textbf{38}(8),  1875--1884 (27~Aug 2019). \doi{10.1109/tmi.2019.2902044}

\bibitem{Litrico2024-wd}
Litrico, M., Guarnera, F., Giuffirda, V., Ravì, D., Battiato, S.: {TADM}: Temporally-aware diffusion model for neurodegenerative progression on brain {MRI}. arXiv [eess.IV]  (18~Jun 2024)

\bibitem{Menze2015-gq}
Menze, B.H., Jakab, A., Bauer, S., Kalpathy-Cramer, J., Farahani, K., Kirby, J., et~al.: The multimodal brain tumor image segmentation benchmark ({BRATS}). IEEE Trans. Med. Imaging  \textbf{34}(10),  1993--2024 (Oct 2015). \doi{10.1109/TMI.2014.2377694}

\bibitem{Metzcar2024-it}
Metzcar, J., Jutzeler, C.R., Macklin, P., Köhn-Luque, A., Brüningk, S.C.: A review of mechanistic learning in mathematical oncology. Front. Immunol.  \textbf{15},  1363144 (12~Mar 2024). \doi{10.3389/fimmu.2024.1363144}

\bibitem{Nichol2021-sc}
Nichol, A., Dhariwal, P.: Improved denoising diffusion probabilistic models  (18~Feb 2021)

\bibitem{Pati2020-py}
Pati, S., Singh, A., Rathore, S., Gastounioti, A., Bergman, M., Ngo, P., Ha, S.M., Bounias, D., Minock, J., Murphy, G., Li, H., Bhattarai, A., Wolf, A., Sridaran, P., Kalarot, R., Akbari, H., Sotiras, A., Thakur, S.P., Verma, R., Shinohara, R.T., Yushkevich, P., Fan, Y., Kontos, D., Davatzikos, C., Bakas, S.: The cancer imaging phenomics toolkit ({CaPTk}): Technical overview. Brainlesion  \textbf{11993},  380--394 (19~May 2020). \doi{10.1007/978-3-030-46643-5\_38}

\bibitem{Peng2022-er}
Peng, W., Adeli, E., Zhao, Q., Pohl, K.M.: Generating realistic {3D} brain {MRIs} using a conditional diffusion probabilistic model  (15~Dec 2022)

\bibitem{Pinaya2022-se}
Pinaya, W.H.L., Tudosiu, P.D., Dafflon, J., da~Costa, P.F., Fernandez, V., Nachev, P., Ourselin, S., Cardoso, M.J.: Brain imaging generation with latent diffusion models. arXiv [eess.IV]  (15~Sep 2022)

\bibitem{Ronneberger2015-se}
Ronneberger, O., Fischer, P., Brox, T.: {U}-net: Convolutional networks for biomedical image segmentation. In: Medical Image Computing and Computer-Assisted Intervention – MICCAI 2015. pp. 234--241. Springer International Publishing (2015). \doi{10.1007/978-3-319-24574-4\_28}

\bibitem{Song2020-vt}
Song, J., Meng, C., Ermon, S.: Denoising diffusion implicit models  (6~Oct 2020)

\bibitem{2020SciPy-NMeth}
Virtanen, P., Gommers, R., Oliphant, T.E., Haberland, M., Reddy, T., Cournapeau, D., Burovski, E., Peterson, P., Weckesser, W., Bright, J., {van der Walt}, S.J., Brett, M., Wilson, J., Millman, K.J., Mayorov, N., Nelson, A.R.J., Jones, E., Kern, R., Larson, E., Carey, C.J., Polat, {\.I}., Feng, Y., Moore, E.W., {VanderPlas}, J., Laxalde, D., Perktold, J., Cimrman, R., Henriksen, I., Quintero, E.A., Harris, C.R., Archibald, A.M., Ribeiro, A.H., Pedregosa, F., {van Mulbregt}, P., {SciPy 1.0 Contributors}: {{SciPy} 1.0: Fundamental Algorithms for Scientific Computing in Python}. Nature Methods  \textbf{17},  261--272 (2020). \doi{10.1038/s41592-019-0686-2}

\bibitem{Wen2010-ex}
Wen, P.Y., Macdonald, D.R., Reardon, D.A., Cloughesy, T.F., Sorensen, A.G., Galanis, E., DeGroot, J., Wick, W., Gilbert, M.R., Lassman, A.B., Tsien, C., Mikkelsen, T., Wong, E.T., Chamberlain, M.C., Stupp, R., Lamborn, K.R., Vogelbaum, M.A., van~den Bent, M.J., Chang, S.M.: Updated response assessment criteria for high-grade gliomas: Response assessment in neuro-oncology working group. J. Clin. Oncol.  \textbf{28}(11),  1963--1972 (10~Apr 2010). \doi{10.1200/JCO.2009.26.3541}

\bibitem{Wolleb2022-dx}
Wolleb, J., Sandkühler, R., Bieder, F., Cattin, P.C.: The swiss army knife for image-to-image translation: Multi-task diffusion models  (6~Apr 2022)

\bibitem{Yankeelov2013-ht}
Yankeelov, T.E., Atuegwu, N., Hormuth, D., Weis, J.A., Barnes, S.L., Miga, M.I., Rericha, E.C., Quaranta, V.: Clinically relevant modeling of tumor growth and treatment response. Sci. Transl. Med.  \textbf{5}(187),  187ps9 (29~May 2013). \doi{10.1126/scitranslmed.3005686}

\bibitem{Zheng2025-ta}
Zheng, D., Preuss, K., Milano, M.T., He, X., Gou, L., Shi, Y., Marples, B., Wan, R., Yu, H., Du, H., Zhang, C.: Mathematical modeling in radiotherapy for cancer: a comprehensive narrative review. Radiat. Oncol.  \textbf{20}(1), ~49 (4~Apr 2025). \doi{10.1186/s13014-025-02626-7}

\end{thebibliography}
%




\end{document}